\documentclass[10pt,twocolumn,letterpaper]{article}

\usepackage{cvpr}

\usepackage{cuted} 

\usepackage{currfile} 

\usepackage{caption} 

\usepackage{multirow}
\usepackage{placeins}

\definecolor{cvprblue}{rgb}{0.21,0.49,0.74}
\usepackage[breaklinks,colorlinks,allcolors=cvprblue]{hyperref}

\title{TransArt3D: From Two States to Articulated 3D Reconstruction with Video Transition Priors}
\author{
Chenhao Guan$^{*}$ \quad Liying Yang$^{*\dagger}$ \quad Xinxing Yu \quad Hao Mo \\
Tianshun Han \quad Renjun Gao \quad Yanyan Liang$^{\dagger}$ \\[4pt]
Macau University of Science and Technology \\[2pt]
{\normalsize $^{*}$Equal contribution \qquad $^{\dagger}$Corresponding authors} \\[2pt]
{\tt\small chengu323@gmail.com \qquad yyliang@must.edu.mo}
}

\begin{document}
\maketitle
\begin{strip}
\centering
\includegraphics[width=\textwidth]{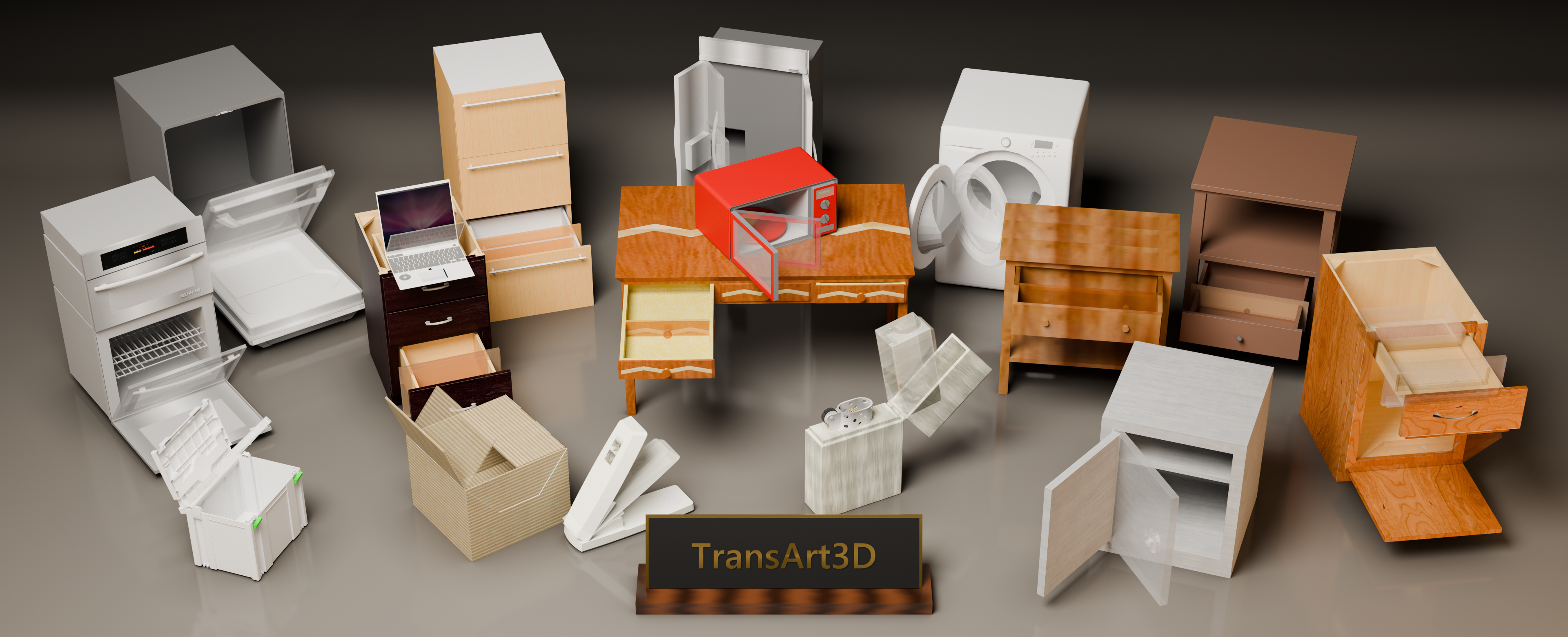}
\captionsetup{hypcap=false}
\captionof{figure}{We present TransArt3D, which reconstructs interactable articulated 3D assets from two endpoint images of an object by translating a video transition prior into explicit parts and a joint. Each asset is shown posed along its recovered joint, with the movable part rendered in transparency at an intermediate state. TransArt3D generalizes across object categories and joint types without any articulated training data.}
\captionsetup{hypcap=true}
\label{fig:teaser}%
\end{strip}

\begin{abstract}
Reconstructing interactable articulated assets from a few images is key to robotics and simulation, yet highly underconstrained. Existing optimization-based methods fit an articulated asset under a static 3D prior that scores each state independently, so per-state reconstruction errors leak into the estimated moving part and joint as false motion. Video diffusion models capture how objects move between states, but only implicitly in 2D. To this end, we present TransArt3D, an optimization-based framework that reconstructs an articulated 3D asset from two endpoint images by translating this video transition prior into explicit part structure and joint kinematics. It first lifts frames generated between the endpoints with a transition-coupled 3D consensus that applies the static 3D prior jointly across frames, stabilizing unchanging regions and initializing the parts and joint. A rectified-flow kinematic refinement then lets the video prior supervise a 3D motion field over the whole transition, while part assignment and a shared joint constrain the field throughout optimization, keeping per-state errors out of the articulation. Finally, a motion-aware canonical appearance fusion uses the recovered articulation to map both endpoint images onto one canonical surface, texturing regions that only one state reveals. On PartNet-Mobility, TransArt3D outperforms methods given the same two endpoints in geometry and joint accuracy, and surpasses methods given six observed states in articulated-state geometry, joint localization and appearance.
\end{abstract}
\section{Introduction}
\label{sec:intro}
Articulated 3D assets are central to robotics and simulation, where digital replicas support interaction and reconfiguration.
Building such an asset requires the geometry and appearance of the object, together with its movable parts and the joints that drive them.
A few images of the object at different states show only a small part of this articulation.
The reconstruction is therefore highly underconstrained.

\begin{figure}[t]
    \centering
    \includegraphics[width=\columnwidth]{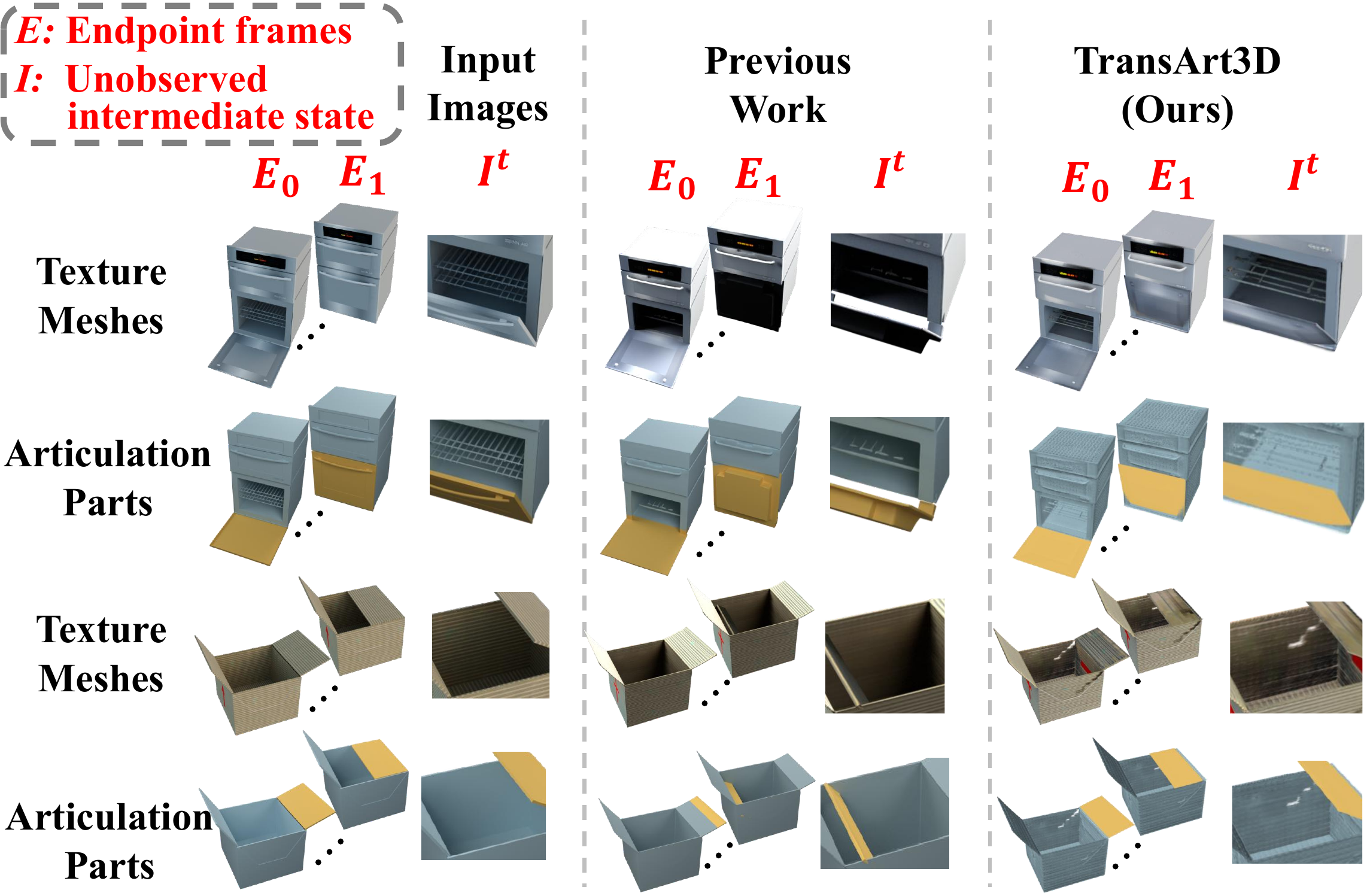}
    \caption{Per-state errors become false motion. FreeArt3D~\cite{freeart3d} scores each state independently, so reconstruction errors of single states enter the moving part and joint and leave the parts and the unobserved intermediate state $I^t$ inconsistent. TransArt3D fits one joint to the whole generated transition, so parts and poses stay consistent across states.}
    \label{fig:freeart3d_failure}
\end{figure}

Learned methods predict parts and joints from a few images~\cite{art,larm,pact} or from a static mesh~\cite{li2026particulate}.
Others generate a part layout and retrieve part meshes~\cite{singapo,dipo,articulateanything}.
Their outputs are bounded by articulated training data, which is orders of magnitude smaller than static 3D and video data.
Optimization-based methods fit an articulated asset to the observed states under a pretrained static 3D prior.
FreeArt3D~\cite{freeart3d} scores each posed state with this prior independently.
Nothing in this score ties the states together, so per-state reconstruction errors leak into the estimated moving part and joint as false motion (Fig.~\ref{fig:freeart3d_failure}).

Video diffusion models offer a complementary source of motion knowledge.
Trained on large video collections, they model how objects move between states.
In-betweening models generate the transition between two given frames~\cite{wan}.
This knowledge is implicit in 2D.
A generated video does not specify which 3D elements move together or which joint drives them.
CHORD~\cite{chord} distills a video prior into a free-form 3D motion field.
This field can explain the transition with any motion, so it yields no part and no joint.
Lifting each generated frame to 3D on its own brings back the per-state errors of the static prior, which a joint fitted to the frames reads as motion.

Both shortcomings point to the same remedy.
Per-state errors change from frame to frame, and true articulation is one rigid motion shared by all frames.
A joint that must explain every frame can therefore fit only the shared motion.
Fitting parts and one joint to the whole transition thus turns the video prior into articulation.

We therefore present TransArt3D, an optimization-based framework that reconstructs an articulated 3D asset from two unposed endpoint images.
It translates the video transition prior into explicit part structure and joint kinematics by letting the prior supervise a 3D motion field over the whole transition.
A part assignment and one shared joint constrain this field throughout optimization.

\begin{figure*}[t]
    \centering
    \includegraphics[width=\textwidth]{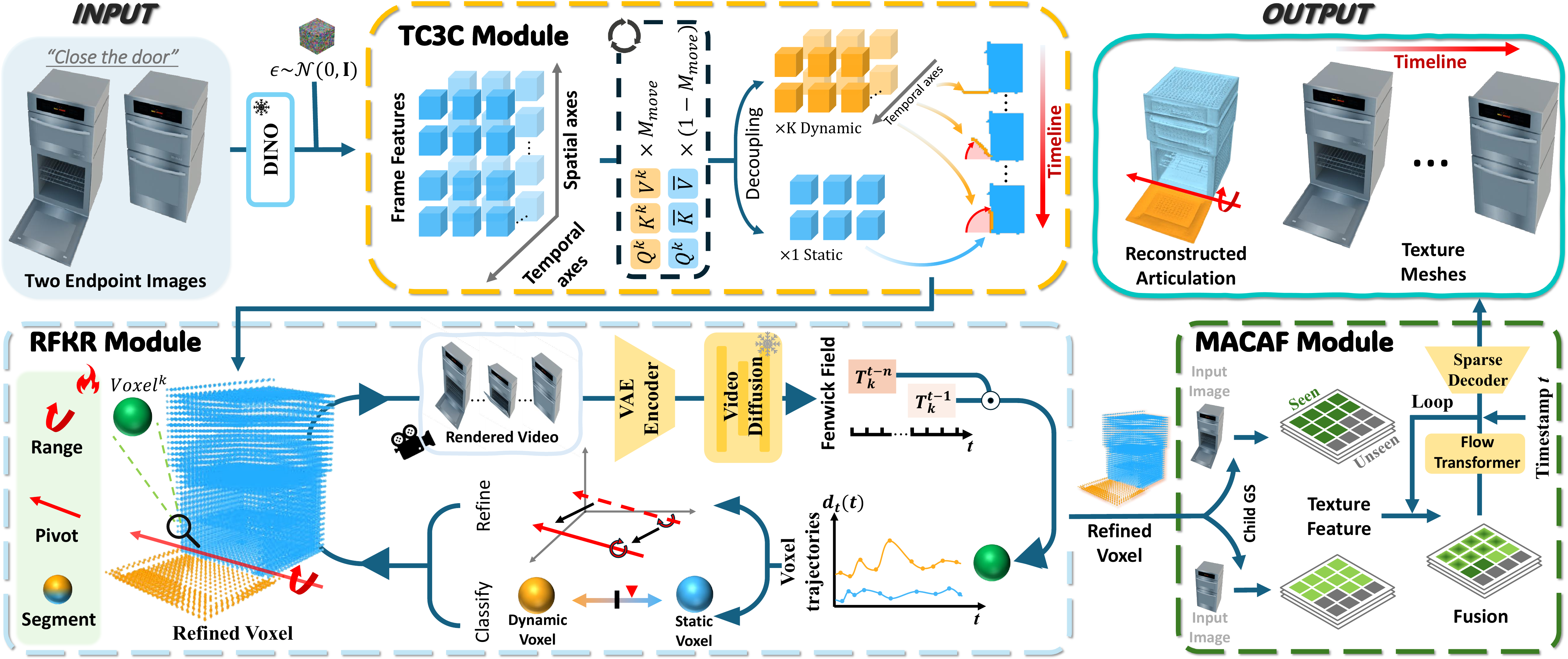}
    \caption{\textbf{Overview of TransArt3D.} A frozen video model generates the transition between the two endpoint images. Transition-Coupled 3D Consensus (TC3C) lifts the generated frames to 3D jointly, which yields an initial part assignment and joint. Rectified-Flow Kinematic Refinement (RFKR) then lets the video prior correct a 3D motion field over the whole transition, while one shared joint constrains the field and is refined by it. The recovered articulation finally lets Motion-Aware Canonical Appearance Fusion (MACAF) map both endpoint images onto one canonical surface. The output is a textured articulated asset posed along the recovered joint.}
    \label{fig:pipeline}
\end{figure*}

TransArt3D couples the states at every step in geometry, in motion and in appearance (Fig.~\ref{fig:pipeline}).
A transition-coupled 3D consensus lifts the frames generated between the endpoints to 3D.
It applies the static 3D prior jointly across frames.
Unchanging regions thus agree across states, and the changing region yields the initial parts and joint.
This initialization is fitted to the lifted frames and inherits their errors.
A rectified-flow kinematic refinement therefore lets the video prior correct the motion over the whole transition.
It builds on the control-point motion field and video distillation of CHORD~\cite{chord} and splits this field into a static base and a movable part tied to the shared joint.
The corrected field in turn refines the part assignment and the joint.
The refined articulation then determines which surface each endpoint image observes.
A motion-aware canonical appearance fusion uses it to map both images onto one canonical surface.
Regions that only one state reveals are textured from that state.

Our main contributions are as follows.

\begin{itemize}
    \item We present TransArt3D, an optimization-based framework that reconstructs an articulated 3D asset from two endpoint images by translating a video transition prior into explicit part structure and joint kinematics.
    \item We propose a rectified-flow kinematic refinement in which a part assignment and a shared joint constrain a video-supervised 3D motion field throughout optimization, keeping per-state errors out of the articulation.
    \item We introduce a transition-coupled 3D consensus that applies the static 3D prior jointly across generated frames to initialize the parts and joint, and a motion-aware canonical appearance fusion that textures one canonical surface from both endpoint images.
    \item Experiments on PartNet-Mobility and real-world captures demonstrate that TransArt3D outperforms existing methods in geometry, articulation and appearance, and that two endpoint images suffice to surpass methods given six observed states.
\end{itemize}

\section{Related Work}

\textbf{Articulated Object Reconstruction.}
Optimization-based methods fit parts and joints to multi-view or multi-state observations with neural fields or Gaussians~\cite{articulate_nerf,paoli,articulatedgs,paris,part2gs}.
Most of them rely on many views per state and on calibrated cameras.
Learned methods predict parts and joints from one or a few images~\cite{art,singapo,pact,dipo} or from a static mesh~\cite{li2026particulate}.
LARM~\cite{larm} additionally receives the joint as a condition, and Articulate-Anything~\cite{articulateanything} assembles the asset from a part database.
The outputs of these methods are bounded by articulated training data and part libraries.
FreeArt3D~\cite{freeart3d} lifts this bound by fitting canonical geometry, texture and joint parameters under a frozen static $3$D prior.
This prior scores one posed state per query, so nothing in it decides which points move together or which joint explains the differences between states.
TransArt3D keeps the frozen static prior and obtains this cross-state evidence from a video transition prior.

\textbf{Appearance of Articulated Objects.}
Texturing methods paint a fixed geometry from text or images~\cite{vcdtexture,flexpainter,texgen,paint3d}, and articulated generation methods synthesize appearance together with structure~\cite{artilatent,dreamart}.
Surfaces that only one state reveals receive no observation in either setting.
TransArt3D uses the recovered articulation to route each endpoint image to the canonical surface it observes.

\textbf{Video Priors for Articulation.}
Video diffusion models provide a prior over plausible transitions between states~\cite{wan}.
ATOP personalizes this prior per object~\cite{atop}, DreamArt fine-tunes it on articulated data~\cite{dreamart}, and SP4D predicts RGB and kinematic-part videos~\cite{sp4d}.
AKD distills the prior into a supplied rig~\cite{akd}.
In-2-4D and TwoSquared lift image pairs to $4$D with free-form deformation and recover neither parts nor a joint~\cite{in24d,twosquared}.
CHORD distills the prior into a control-point motion field that has the same limit~\cite{chord}.
TransArt3D builds on the field and distillation of CHORD and lets a part assignment and one shared joint constrain the field throughout optimization, which turns the corrected motion into explicit articulation.

\section{Method}

\subsection{Overview}
\label{sec:overview}

Given unposed images $I_0$ and $I_1$ of an articulated object at its start and end states and a text motion prompt $c$, we reconstruct a textured articulated 3D asset.
We assume a single joint and take the start state as canonical.
A part assignment splits canonical points into a static base $\mathcal{B}$ and a movable part $\mathcal{M}$.
We infer the joint type and its parameters $\mathcal{J}$.
A revolute joint has $\mathcal{J} = (\mathbf{a}, \mathbf{o}, \rho)$ with unit axis $\mathbf{a}$, pivot $\mathbf{o}$ on the axis and angular range $\rho$.
A prismatic joint has $\mathcal{J} = (\mathbf{a}, \rho)$ with unit axis $\mathbf{a}$ and translation range $\rho$.
Each state has a joint state $q \in [0, \rho]$, with $q = 0$ at $I_0$ and $q = \rho$ at $I_1$.
The transform $\mathbf{T}_q$ rotates $\mathcal{M}$ by $q$ about the axis through $\mathbf{o}$, or translates it by $q$ along $\mathbf{a}$.

Figure~\ref{fig:pipeline} shows an overview of TransArt3D.
A frozen video diffusion model conditioned on $I_0$, $I_1$ and $c$ guides a 3D motion field, which the part assignment and joint constrain.
TC3C (Sec.~\ref{sec:TC3C}) lifts frames generated by this model to 3D with TRELLIS~\cite{trellis} and initializes the canonical support, part assignment and joint.
RFKR (Sec.~\ref{sec:RFKR}) jointly refines the motion field, part assignment and joint.
MACAF (Sec.~\ref{sec:MACAF}) uses the recovered part assignment and joint to fuse $I_0$ and $I_1$ into one canonical appearance.

\subsection{Preliminaries}
\label{sec:prelim}

\noindent\textbf{TRELLIS.}
TRELLIS~\cite{trellis} generates a 3D asset from an image in two stages.
Each stage runs a rectified-flow transformer conditioned on DINOv2~\cite{oquab2023dinov2} features of the image.
Rectified flow interpolates between data $\mathbf{z}$ and Gaussian noise $\boldsymbol{\epsilon}$ at flow time $\tau \in [0, 1]$ as
\begin{equation}
\mathbf{z}_\tau = (1 - \tau)\,\mathbf{z} + \tau\,\boldsymbol{\epsilon},
\label{eq:rf}
\end{equation}
and the transformer predicts the velocity $\boldsymbol{\epsilon} - \mathbf{z}$.
Each sampling step also gives a clean estimate $\hat{\mathbf{z}}$ from the predicted velocity.
The first stage samples a $16^3$ latent grid of tokens indexed by $j$ and decodes it into a $64^3$ occupancy grid with occupied voxels $\mathbf{x}_i$.
The second stage attaches a latent $\mathbf{z}_i$ to each occupied voxel, forming the structured latent $\mathcal{Z} = \{(\mathbf{x}_i, \mathbf{z}_i)\}$.
The structured-latent encoder computes $\mathcal{Z}$ from DINOv2 features aggregated over rendered views, and its decoders map $\mathcal{Z}$ to 3D Gaussians or a mesh.

\noindent\textbf{CHORD.}
CHORD~\cite{chord} distills motion from a rectified-flow video model into a control-point motion field.
Each control point $k$ has a center $\boldsymbol{\mu}_k$ and a rigid transform $(\mathbf{R}_k(t), \mathbf{b}_k(t))$ at every frame $t$.
A Fenwick tree stores these transforms over frame ranges, so nearby frames share parameters.
A canonical point $\mathbf{x}$ moves by
\begin{equation}
\mathbf{d}(\mathbf{x}, t; \theta) = \sum_{k \in \mathcal{N}(\mathbf{x})} w_k(\mathbf{x}) \big[(\mathbf{R}_k(t) - \mathbf{I})(\mathbf{x} - \boldsymbol{\mu}_k) + \mathbf{b}_k(t)\big],
\label{eq:field}
\end{equation}
where $\mathcal{N}(\mathbf{x})$ holds its nearest control points, $w_k(\mathbf{x})$ are normalized Gaussian skinning weights and $\theta$ collects all transforms.
At each iteration, canonical 3D Gaussians deformed by this field are rendered from sampled cameras and encoded into a video latent $\mathbf{z}^{\mathrm{v}}$, which is noised to $\mathbf{z}^{\mathrm{v}}_\tau$ by Eq.~\eqref{eq:rf}.
With the frozen velocity predictor $\hat{\mathbf{v}}_\psi$ and its condition $\mathbf{y}$, the video residual and the W-RFSDS update are
\begin{equation}
\begin{aligned}
\mathbf{r}_\psi &= \hat{\mathbf{v}}_\psi(\mathbf{z}^{\mathrm{v}}_\tau; \tau, \mathbf{y}) - (\boldsymbol{\epsilon} - \mathbf{z}^{\mathrm{v}}),\\
\nabla_\theta \mathcal{L}_{\text{W-RFSDS}} &= \mathbb{E}_{\boldsymbol{\epsilon}}\Big[\mathbf{r}_\psi \frac{\partial \mathbf{z}^{\mathrm{v}}}{\partial \theta}\Big],
\end{aligned}
\label{eq:wrfsds}
\end{equation}
where $\tau$ is set at each iteration by a descending annealing schedule.
\subsection{Transition-Coupled 3D Consensus}
\label{sec:TC3C}

TC3C lifts frames of the generated transition to 3D and initializes the canonical support, part assignment and joint.
Lifting each frame with TRELLIS independently gives a plausible shape per state, yet these shapes differ even on the static base.
These per-state reconstruction errors then leak into the fitted part and joint as false motion.
TC3C therefore samples all states jointly, so that unchanging regions agree across states, and fits the joint and part assignment to the coupled result.

\noindent\textbf{Cross-state coupling.}
Intuitively, structure that persists through the transition should agree across states, while the region swept by $\mathcal{M}$ should keep state-specific structure.
We sample $K$ frames uniformly from the transition, with frames $1$ and $K$ at $I_0$ and $I_1$, and condition the first TRELLIS stage on each of them.
All states share one noise draw, so token $j$ refers to the same location in every state.
Sampling runs in two passes.
The first pass mixes the latent of each state with those of the two endpoints during early steps.
The second pass re-noises this result and replaces self-attention with a gated fusion of a state-specific and a cross-state response,
\begin{equation}
\mathbf{h}_{k,j} = g_j\,\mathbf{A}_k(j) + (1 - g_j)\,\bar{\mathbf{A}}_k(j),
\label{eq:gate}
\end{equation}
where $\mathbf{A}_k(j)$ attends from token $j$ to the tokens of state $k$, and $\bar{\mathbf{A}}_k(j)$ uses the same query with keys and values averaged over states.
The motion-selective gate $g_j$ grows with the cross-state disagreement of the first-pass occupancies and of the current clean estimates $\hat{\mathbf{z}}$.
It shrinks on confident persistent structure and stays below $1$.
Decoding gives binary occupancies $O_k$ of the $K$ states on a shared $64^3$ grid.

\noindent\textbf{Articulation initialization.}
Removing the voxels occupied in nearly all states leaves the change region of each state.
A translation preserves the projection of the movable part onto the plane normal to its axis, whereas a rotation generally does not.
We compare these projections across states for each axis direction of the canonical frame, which selects the joint type and the axis of a prismatic joint.
A grid search followed by sub-voxel refinement then finds the joint state $q_k$ of each state $k$.
For a revolute joint, it also selects the axis among these directions and finds the pivot $\mathbf{o}$.
These parameters maximize the overlap between the largest connected part of the canonical change region, moved by $\mathbf{T}_{q_k}$, and the voxels of $O_k$ not occupied in every state.
We fix $q_1 = 0$ and orient $\mathbf{a}$ so that $q_K \ge 0$.
This gives the initial joint $\mathcal{J}_0$ with $\rho = q_K$.
Given $\mathcal{J}_0$, each voxel of the canonical support $\mathcal{U} = O_1$ either stays fixed or moves with the joint.
We label it with the explanation that fewer states contradict.
Smoothness and collision terms regularize this labeling, which gives the initial base $\mathcal{B}_0$ and movable part $\mathcal{M}_0$.
The second TRELLIS stage, conditioned on $I_0$, generates a structured latent on $\mathcal{U}$, which we decode into canonical 3D Gaussians.

\subsection{Rectified-Flow Kinematic Refinement}
\label{sec:RFKR}

RFKR refines the motion field, part assignment and joint together.
The initialization from TC3C is fitted to lifted frames and inherits their errors.
The video residual of Eq.~\eqref{eq:wrfsds} acts on rendered pixels and does not indicate which voxels move together.
RFKR therefore lets the video prior supervise a 3D motion field over the whole transition, while the part assignment and a shared joint constrain this field throughout optimization.

\noindent\textbf{Formulation.}
The field of Eq.~\eqref{eq:field} acts on the canonical voxels, with $\mathbf{d}_i(t) = \mathbf{d}(\mathbf{x}_i, t; \theta)$ at frame $t \in \{0, \dots, F-1\}$, and each canonical Gaussian follows the displacement of its voxel.
Frame $t$ corresponds to the linear joint state $q_t = \rho\, t/(F-1)$.
We initialize $\rho$ and $\mathbf{o}$ from $\mathcal{J}_0$ and seed $\theta$ in closed form to reproduce the joint motion at these states.
The control points inside $\mathcal{M}_0$ form the movable group, and the rest form the base group.
Summing their contributions separately splits $\mathbf{d}_i(t)$ into $\mathbf{d}^{\mathcal{M}}_i(t)$ and $\mathbf{d}^{\mathcal{B}}_i(t)$, and $\alpha_i$ denotes the total skinning weight of the movable group at $\mathbf{x}_i$.
We optimize $\theta$, the range $\rho$ and, for a revolute joint, the offset of $\mathbf{o}$ within the plane normal to $\mathbf{a}$.
The part assignment is also optimized within a one-voxel band $\mathcal{S}_0$ around the boundary between $\mathcal{B}_0$ and $\mathcal{M}_0$.
There, $\pi_i = \sigma\big((\ell_i + \Delta(\mathbf{x}_i))/\kappa\big)$ is the probability that $\mathbf{x}_i$ belongs to $\mathcal{M}$, with temperature $\kappa$.
Here $\ell_i$ is the prior logit from TC3C, and $\Delta$ is a zero-initialized coordinate MLP.
Outside $\mathcal{S}_0$, $\pi_i$ keeps its binary label from TC3C.

\noindent\textbf{Video guidance.}
At each iteration, the deformed Gaussians are rendered over all $F$ frames from sampled cameras.
The condition $\mathbf{y}$ holds $c$ and two endpoint frames, rendered from the same cameras with the initialized asset posed by $\mathcal{J}_0$ at its endpoints.
These endpoint frames do not change with $\theta$.
They determine the first and last latent frames, which we exclude from $\mathbf{r}_\psi$.
The update of Eq.~\eqref{eq:wrfsds} reaches $\theta$ through the rendered transition, and Eq.~\eqref{eq:coupling} below carries the corrected trajectories to the joint and the part assignment.

\noindent\textbf{Kinematic coupling.}
The joint moves voxel $\mathbf{x}_i$ by $\hat{\mathbf{d}}_i(t) = \mathbf{T}_{q_t}(\mathbf{x}_i) - \mathbf{x}_i$ at frame $t$.
We ask the movable group to follow this motion and the base group to stay still,
\begin{equation}
\begin{aligned}
E_{\mathrm{dyn}} &= \sum_{t \ge 1} \sum_i \pi_i \big\|\mathbf{d}^{\mathcal{M}}_i(t) - \alpha_i \hat{\mathbf{d}}_i(t)\big\|_2^2,\\
E_{\mathrm{stat}} &= \frac{1}{F-1} \sum_{t \ge 1} \sum_i (1 - \pi_i) \big\|\mathbf{d}^{\mathcal{B}}_i(t)\big\|_2^2.
\end{aligned}
\label{eq:coupling}
\end{equation}
The joint receives gradients only through $E_{\mathrm{dyn}}$, which refines the pivot and range from the corrected motion.
$E_{\mathrm{dyn}}$ sums over all frames, so one joint explains the whole corrected trajectory.
Inside $\mathcal{S}_0$, the relative cost of the two terms moves $\pi_i$ toward the explanation that better fits the trajectory of $\mathbf{x}_i$.
A seam-cut rigidity term further aligns the boundary with discontinuities of the corrected field,
\begin{equation}
E_{\mathrm{seam}} = \sum_{(i,j)} \big[\pi_i \pi_j + (1 - \pi_i)(1 - \pi_j)\big]\, e_{ij},
\label{eq:seam}
\end{equation}
where $(i, j)$ runs over a fixed nearest-neighbor graph on the canonical voxels.
The cost $e_{ij}$ measures how far the relative motion of $\mathbf{x}_j$ departs from the local rotation at $\mathbf{x}_i$ over all frames.
A large $e_{ij}$ lowers the cost of a label change between $\mathbf{x}_i$ and $\mathbf{x}_j$.

\noindent\textbf{Optimization.}
Each iteration minimizes
\begin{equation}
\mathcal{L} = \mathcal{L}_{\text{W-RFSDS}} + \lambda_c E_{\mathrm{dyn}} + \lambda_S E_{\mathrm{stat}} + \lambda_{\mathrm{seam}} E_{\mathrm{seam}} + E_{\mathrm{reg}},
\label{eq:rfkr_total}
\end{equation}
where $E_{\mathrm{reg}}$ holds the temporal smoothness term of CHORD and a neighbor smoothness term on $\pi$.
The weight $\lambda_c$ starts small and grows during optimization.
The joint thus constrains the field from the first iteration and binds it more tightly later.
Thresholding $\pi$ gives the refined base $\mathcal{B}$ and movable part $\mathcal{M}$, which MACAF uses together with the refined joint $\mathcal{J}$.

\subsection{Motion-Aware Canonical Appearance Fusion}
\label{sec:MACAF}

\begin{table*}[!t]
    \centering
    \begingroup
    \small
    \setlength{\tabcolsep}{2.7pt}
    \begin{tabular*}{\textwidth}{@{\extracolsep{\fill}}cl*{11}{c}@{}}
        \toprule
        & \multirow{2}{*}{Method}
        & \multicolumn{2}{c}{$d_{\mathrm{gIoU}}\downarrow$}
        & \multicolumn{2}{c}{$d_{\mathrm{cDist}}\downarrow$}
        & \multicolumn{2}{c}{$d_{\mathrm{CD}}\downarrow$}
        & \multirow{2}{*}{AOR $\downarrow$}
        & \multirow{2}{*}{CLIP $\uparrow$}
        & \multirow{2}{*}{LPIPS $\downarrow$}
        & \multirow{2}{*}{Range $\downarrow$}
        & \multirow{2}{*}{Pivot $\downarrow$} \\
        \cmidrule(lr){3-4}\cmidrule(lr){5-6}\cmidrule(lr){7-8}
        & & RS & AS & RS & AS & RS & AS & & & & & \\
        \midrule
        \multirow{4}{*}{\rotatebox{90}{\shortstack{Multi-\\state}}}
        & FreeArt3D~\cite{freeart3d} & \textbf{0.466} & \textbf{0.376} & \textbf{0.073} & \textbf{0.068} & \textbf{0.052} & \textbf{0.047} & \underline{0.035} & \textbf{0.886} & \underline{0.064} & \textbf{0.163} & \textbf{0.134} \\
        & PARIS~\cite{paris} & 0.871 & 0.896 & \underline{0.201} & \underline{0.217} & 0.221 & 0.184 & 0.053 & 0.736 & 0.102 & 0.871 & \underline{0.315} \\
        & SINGAPO$^\dagger$~\cite{singapo} & 0.525 & 0.518 & 0.147 & 0.196 & 0.067 & 0.029 & 0.005 & 0.826 & 0.088 & 0.853 & 0.248 \\
        & Arti-Anything~\cite{articulateanything} & \underline{0.778} & \underline{0.758} & 0.217 & 0.284 & \underline{0.103} & \underline{0.174} & \textbf{0.024} & \underline{0.829} & \textbf{0.026} & \underline{0.452} & 0.447 \\
        \midrule
        \multirow{6}{*}{\rotatebox{90}{\shortstack{Dual-\\state}}}
        & FreeArt3D~\cite{freeart3d} & 1.026 & 0.948 & 0.228 & \underline{0.234} & 0.269 & 0.296 & 0.093 & \underline{0.876} & 0.072 & 1.051 & 0.352 \\
        & PARIS~\cite{paris} & 0.895 & 0.947 & 0.204 & 0.243 & 0.235 & 0.286 & 0.077 & 0.718 & 0.083 & 1.224 & \underline{0.254} \\
        & SINGAPO$^\dagger$~\cite{singapo} & 0.531 & 0.545 & 0.153 & 0.197 & 0.075 & 0.024 & 0.008 & 0.825 & 0.085 & 0.908 & 0.293 \\
        & Arti-Anything~\cite{articulateanything} & 0.705 & 0.787 & 0.233 & 0.297 & \underline{0.117} & \underline{0.194} & 0.026 & 0.830 & \textbf{0.026} & \underline{0.454} & 0.531 \\
        & DIPO~\cite{dipo} & 1.145 & 1.045 & 0.227 & 0.265 & 0.310 & 0.354 & \textbf{0.006} & 0.833 & 0.077 & 0.484 & 0.431 \\
        & LARM ~\cite{larm} & \underline{0.704} & \underline{0.740} & \underline{0.169} & 0.241 & 0.205 & 0.311 & 0.094 & 0.779 & 0.081 & 0.671 & 1.159 \\
        & \textbf{TransArt3D (Ours)} & \textbf{0.473} & \textbf{0.357} & \textbf{0.078} & \textbf{0.063} & \textbf{0.080} & \textbf{0.041} & \underline{0.025} & \textbf{0.908} & \underline{0.052} & \textbf{0.223} & \textbf{0.077} \\
        \bottomrule
    \end{tabular*}
    \endgroup
    \caption{Quantitative comparison with six observed states in the multi-state block and two observed endpoints in the dual-state block, with both evaluated at six configurations. Within each input block, bold and underlined values denote the best and second-best results among methods covering all 48 objects. The $\dagger$ row reports SINGAPO on 28 valid objects from its seven supported categories and is excluded from ranking.}
    \label{tab:main_comparison}
\end{table*}
MACAF recovers one canonical appearance from $I_0$ and $I_1$.
The structured latent from Sec.~\ref{sec:TC3C} is conditioned on $I_0$ alone, so surfaces exposed by the motion receive no observation.
MACAF uses the recovered part assignment and joint to map both endpoint images onto one canonical surface, texturing regions that only one state reveals.

\noindent\textbf{Articulation-guided routing.}
For each endpoint $n \in \{0, 1\}$, we pose the canonical geometry with $\mathbf{T}_0$ or $\mathbf{T}_\rho$ and align its projection to the foreground of $I_n$ by a scale and a translation.
Depth-aware rasterization of the posed base and movable part then gives the rows of $\mathcal{Z}$ visible in $I_n$.
Let $m_{in} \in \{0, 1\}$ indicate this visibility for row $i$, and let $\mathbf{f}_{in}$ be the DINOv2 feature of $I_n$ at the projection of row $i$.
The routed feature of a row visible in at least one image is
\begin{equation}
\bar{\mathbf{f}}_i = \frac{\sum_n m_{in}\, \mathbf{f}_{in}}{\sum_n m_{in}}.
\label{eq:route}
\end{equation}
The measured set $\mathcal{Q}_{\mathrm{meas}}$ holds the rows that only $I_1$ reveals.
It also holds movable-part rows whose decoded color departs from the observed images.
Rows seen in neither image, or seen in both with disagreeing features, are left to the prior, and the other rows keep their latent from Sec.~\ref{sec:TC3C}.
We complete the routed features with features of rendered views of the current asset.
The structured-latent encoder maps this feature field to an initial latent for $\mathcal{Q}_{\mathrm{meas}}$.

\noindent\textbf{Measurement-guided resampling.}
We noise the initial latent and resample it with the second TRELLIS stage, conditioned on $I_1$ for rows that only $I_1$ reveals and on $I_0$ otherwise.
Kept rows are re-injected at every step.
The clean estimate $\hat{\mathbf{z}}$ of each step, with kept rows restored, is decoded, posed at both endpoints and compared with the images,
\begin{equation}
\mathcal{L}_{\mathrm{meas}} = \sum_n \big\| \Omega_n \odot \big(\Phi_n(\hat{\mathbf{z}}) - G(I_n)\big) \big\|_1,
\label{eq:meas}
\end{equation}
where $\Phi_n$ decodes, poses and renders the appearance at endpoint $n$, $\Omega_n$ is the rendered object mask and $G$ is a low-pass filter, also applied inside $\Phi_n$.
The masked gradient $\mathbb{1}_{\mathcal{Q}_{\mathrm{meas}}} \odot \nabla_{\hat{\mathbf{z}}} \mathcal{L}_{\mathrm{meas}}$ updates the sampling state on $\mathcal{Q}_{\mathrm{meas}}$ only, while the prior completes the rows left to it.
We keep only the sampled appearance, and the geometry, part assignment and joint stay unchanged.
The sampled appearance and these fixed quantities form a textured articulated asset that can be posed at any $q \in [0, \rho]$.
\section{Experiments}
\label{sec:exp}

\subsection{Experimental Setup}
\label{sec:setup}

\noindent\textbf{Dataset.}
We evaluate on 48 objects from twelve PartNet-Mobility~\cite{partnetmobility} categories, four per category.
Each object has one revolute or prismatic joint.

The two endpoint images show the resting state and the fully articulated state.

\noindent\textbf{Metrics.}
Each reconstructed asset is posed at six uniformly spaced joint states.
One evaluator renders and scores every method.
We report part geometry at the resting state (RS) and averaged over the five articulated states (AS).
Part geometry is measured by dgIoU, dcDist and dCD~\cite{singapo,pact}.
AOR measures the interpenetration between the two parts~\cite{singapo,pact}.
Joint accuracy is measured by the normalized range error and the axis-normal pivot displacement.
Appearance is measured by CLIP similarity over five views and by foreground-masked LPIPS.
Full metric definitions are provided in the supplementary material.

\begin{figure*}[!t]
    \centering
    \newlength{\qgap}\setlength{\qgap}{1pt}
    \newlength{\qw}\setlength{\qw}{\dimexpr(\textwidth-9\qgap)/10\relax}
    \newcommand{\qim}[1]{\includegraphics[width=\qw,height=\qw]{Figures/#1}}
    \newcommand{\qhead}[1]{\parbox[c]{\dimexpr2\qw+\qgap\relax}{\centering\footnotesize #1}}
    \newcommand{\qstate}[1]{\parbox[c]{\qw}{\centering\scriptsize #1}}
    \newcommand{\qpair}[2]{\qstate{#1}\hspace{\qgap}\qstate{#2}}
    \newcommand{\qrow}[2][]{%
        \hbox{%
            \qim{\ifx\relax#1\relax#2/LARM_1_normalized\else#1\fi/0.png}\hspace{\qgap}%
            \qim{\ifx\relax#1\relax#2/LARM_1_normalized\else#1\fi/1.png}\hspace{\qgap}%
            \qim{submission/#2/aa/0.png}\hspace{\qgap}%
            \qim{submission/#2/aa/1.png}\hspace{\qgap}%
            \qim{submission/#2/fa3d_2/0.png}\hspace{\qgap}%
            \qim{submission/#2/fa3d_2/1.png}\hspace{\qgap}%
            \qim{submission/#2/ours/0.png}\hspace{\qgap}%
            \qim{submission/#2/ours/1.png}\hspace{\qgap}%
            \qim{submission/#2/gt/0.png}\hspace{\qgap}%
            \qim{submission/#2/gt/1.png}}}
    \noindent\makebox[\textwidth][c]{%
        \vbox{%
            \offinterlineskip
            \hbox{%
                \qhead{LARM (one view)}\hspace{\qgap}%
                \qhead{Articulate-Anything}\hspace{\qgap}%
                \qhead{FreeArt3D (2 states)}\hspace{\qgap}%
                \qhead{\textbf{TransArt3D (Ours)}}\hspace{\qgap}%
                \qhead{Input Two Endpoint Images}}
            \vskip\qgap
            \hbox{%
                \qpair{$t=0.5$}{$t=1$}\hspace{\qgap}%
                \qpair{$t=0.5$}{$t=1$}\hspace{\qgap}%
                \qpair{$t=0.5$}{$t=1$}\hspace{\qgap}%
                \qpair{$t=0.5$}{$t=1$}\hspace{\qgap}%
                \qpair{$t=0.5$}{$t=1$}}
            \vskip\qgap
            \qrow{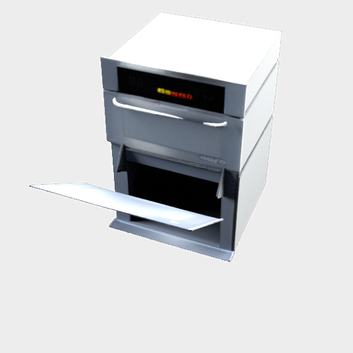}
            \vskip\qgap
            \qrow{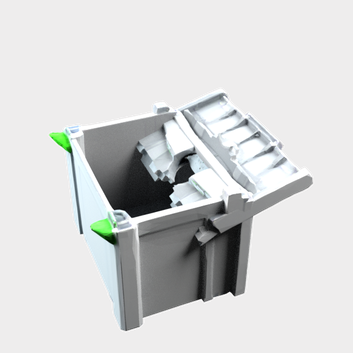}
            \vskip\qgap
            \qrow{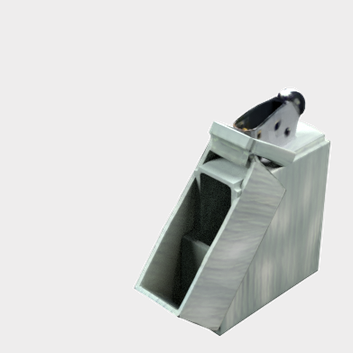}
            \vskip\qgap
            \qrow{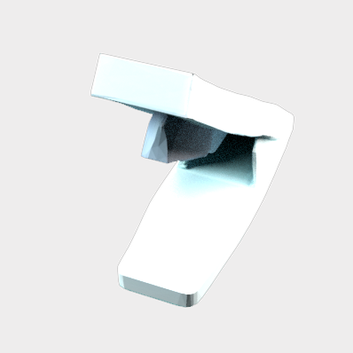}
            \vskip\qgap
            \qrow[submission/45427/larm]{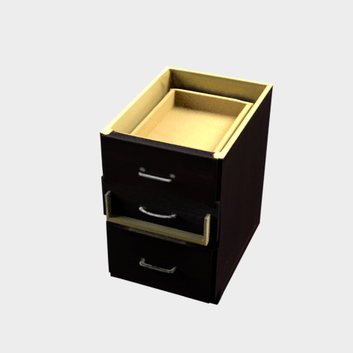}
            \vskip\qgap
            \qrow[submission/20555/larm]{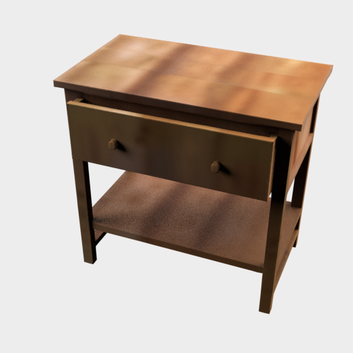}}}
    \caption{Qualitative comparison under the dual-state protocol. Conditioning and adapters for each method are given in the supplementary material, and the shared evaluator renders all outputs with matched cameras.}
    \label{fig:qualitative_comparison}
\end{figure*}
\noindent\textbf{Implementation.}
TransArt3D uses frozen TRELLIS~\cite{trellis} as the static 3D prior and frozen Wan2.2-Fun-A14B-InP~\cite{videoxfun,wan22fun_a14b_inp,wan} as the video transition prior.
It takes the two endpoint images and a text motion prompt as input.
We set $F = 41$ and $K = 6$.
Further settings are provided in the supplementary material.

\subsection{Comparisons with Existing Methods}
\label{sec:comparison}

\noindent\textbf{Baselines.}
We compare with three groups of methods.
LARM~\cite{larm} predicts parts and joints with a feed-forward network.
SINGAPO~\cite{singapo}, DIPO~\cite{dipo} and Articulate-Anything~\cite{articulateanything} generate a part layout and retrieve part meshes.
For a fair comparison, we remove the test objects from their retrieval databases.
PARIS~\cite{paris} and FreeArt3D~\cite{freeart3d} optimize each object.
Under the dual-state protocol, every method receives the same two endpoint images.
Under the multi-state protocol, the baselines receive six state images.
LARM receives one view per state to match our input.
Its released three-view setting is reported in the supplementary material.
SINGAPO supports seven of the twelve categories.
It is evaluated on those 28 objects and marked in Tab.~\ref{tab:main_comparison}.

\begin{table}[t]
    \centering
    \footnotesize
    \setlength{\tabcolsep}{2pt}
    \renewcommand{\arraystretch}{0.96}
    \resizebox{\columnwidth}{!}{%
    \begin{tabular}{@{}l*{6}{c}@{}}
        \toprule
        \multirow{2}{*}{Ablation setting}
        & \multicolumn{3}{c}{AS}
        & \multirow{2}{*}{AOR $\downarrow$}
        & \multirow{2}{*}{Range $\downarrow$}
        & \multirow{2}{*}{Pivot $\downarrow$} \\
        \cmidrule(lr){2-4}
        & $d_{\mathrm{gIoU}}\downarrow$
        & $d_{\mathrm{cDist}}\downarrow$
        & $d_{\mathrm{CD}}\downarrow$
        & & & \\
        \midrule
        \multicolumn{7}{@{}l}{\itshape Cumulative removal} \\
        (a) Full
        & \textbf{0.357} & \textbf{0.063} & \textbf{0.041} & \underline{0.025} & \textbf{0.223} & \textbf{0.077} \\
        (b) w/o Seam Term
        & \underline{0.372} & \underline{0.072} & \underline{0.046} & 0.030 & \underline{0.274} & \underline{0.084} \\
        (c) w/o Joint Refine.
        & 0.436 & 0.092 & 0.076 & \textbf{0.021} & 0.336 & 0.281 \\
        (d) w/o Boundary Refine.
        & 0.504 & 0.112 & 0.108 & 0.036 & 0.566 & 0.283 \\
        (e) w/o Video Guidance
        & 0.518 & 0.125 & 0.110 & 0.034 & 0.613 & 0.282 \\
        (f) w/o Motion Gate
        & 0.672 & 0.151 & 0.166 & 0.029 & 0.687 & 0.319 \\
        (g) w/o Cross-State Coupling
        & 0.969 & 0.244 & 0.304 & 0.108 & 0.741 & 0.427 \\
        \midrule
        \multicolumn{7}{@{}l}{\itshape Component replacement} \\
        (h) Corresp. Init. (GIM-DKM)
        & \textbf{0.352} & \textbf{0.068} & \textbf{0.041} & \textbf{0.038} & \underline{0.218} & \textbf{0.136} \\
        (i) Static 3D Guid. (2 states)
        & \underline{0.385} & 0.218 & 0.274 & 0.073 & 0.928 & 0.224 \\
        (j) Static 3D Guid. (6 states)
        & 0.465 & \underline{0.113} & \underline{0.089} & \underline{0.042} & \textbf{0.175} & \underline{0.148} \\
        \midrule
        \multicolumn{7}{@{}l}{\itshape Post-hoc articulation} \\
        (k) CHORD + Post-hoc Fit
        & 0.502 & 0.061 & 0.092 & 0.001 & 0.359 & 0.156 \\
        \bottomrule
    \end{tabular}}
    \caption{Geometry and articulation ablations on 48 objects. Rows (b) to (g) remove components cumulatively. Rows (h) to (j) each replace one component of Full and keep the rest. Row (k) runs CHORD~\cite{chord} on the ground-truth mesh and fits the parts and joint afterwards. Complete RS and AS results are provided in the supplementary material.}
    \label{tab:ablation_geometry}
\end{table}
\begin{figure}[t]
    \centering
    \begingroup
    \setlength{\tabcolsep}{0.42pt}
    \renewcommand{\arraystretch}{0.9}
    \begin{tabular*}{\columnwidth}{@{\extracolsep{\fill}}*{7}{c}@{}}
        \parbox[c]{0.139\columnwidth}{\centering\small (c)} &
        \parbox[c]{0.139\columnwidth}{\centering\small (d)} &
        \parbox[c]{0.139\columnwidth}{\centering\small (g)} &
        \parbox[c]{0.139\columnwidth}{\centering\small (j)} &
        \parbox[c]{0.139\columnwidth}{\centering\small (k)} &
        \parbox[c]{0.139\columnwidth}{\centering\small Full} &
        \parbox[c]{0.139\columnwidth}{\centering\small GT} \\[-1pt]
        \multicolumn{7}{c}{\small\itshape Textured Meshes} \\[-1pt]
        \includegraphics[width=0.139\columnwidth]{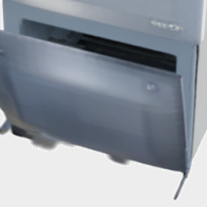} &
        \includegraphics[width=0.139\columnwidth]{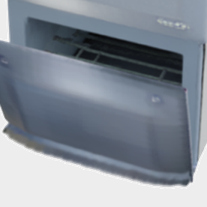} &
        \includegraphics[width=0.139\columnwidth]{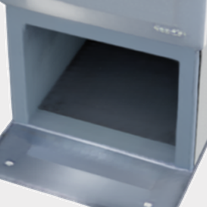} &
        \includegraphics[width=0.139\columnwidth]{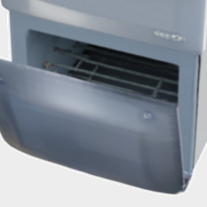} &
        \includegraphics[width=0.139\columnwidth]{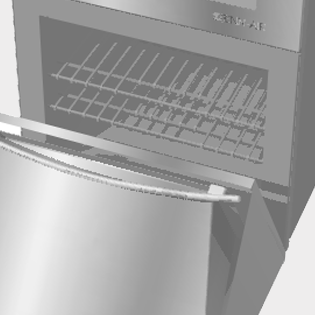} &
        \includegraphics[width=0.139\columnwidth]{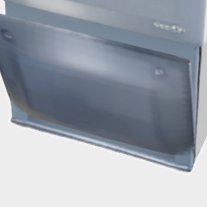} &
        \includegraphics[width=0.139\columnwidth]{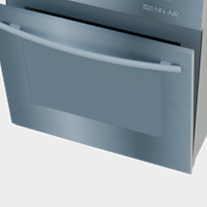} \\
        \multicolumn{7}{c}{\small\itshape Reconstructed Articulation} \\[-1pt]
        \includegraphics[width=0.139\columnwidth]{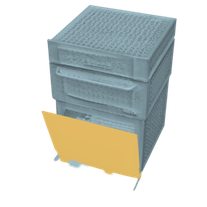} &
        \includegraphics[width=0.139\columnwidth]{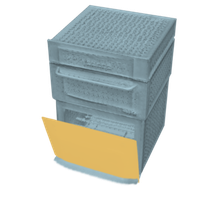} &
        \includegraphics[width=0.139\columnwidth]{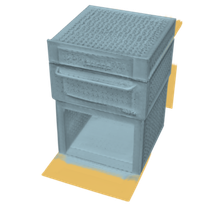} &
        \includegraphics[width=0.139\columnwidth]{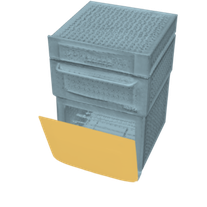} &
        \includegraphics[width=0.139\columnwidth]{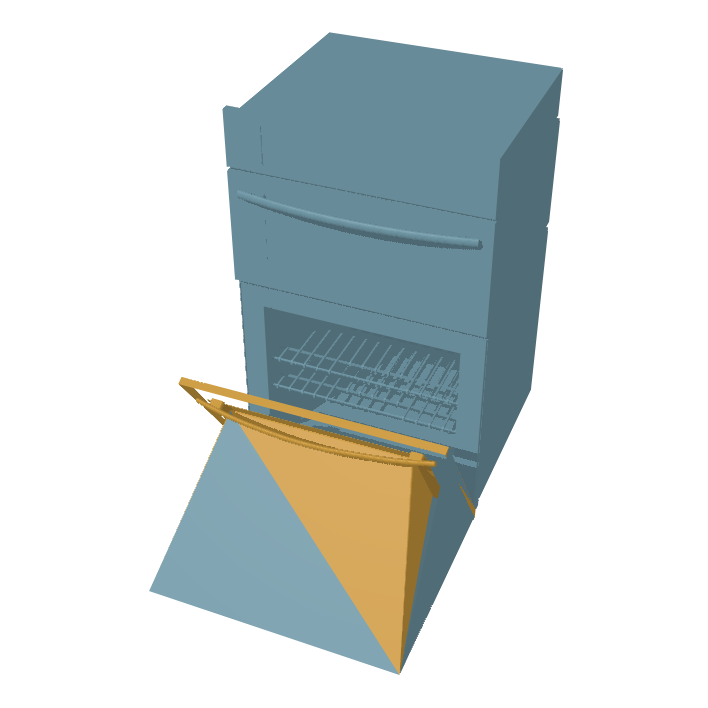} &
        \includegraphics[width=0.139\columnwidth]{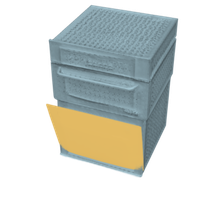} &
        \includegraphics[width=0.139\columnwidth]{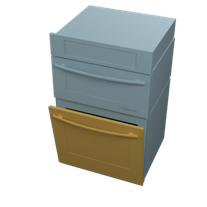}
    \end{tabular*}
    \endgroup
    \caption{Geometry and articulation ablations from Tab.~\ref{tab:ablation_geometry}. Each column shows the textured mesh and the part assignment with the joint.}
    \label{fig:geometry_ablation_viz}
\end{figure}

\noindent\textbf{Quantitative Results.}
Tab.~\ref{tab:main_comparison} shows that TransArt3D achieves the best part geometry and joint accuracy under the dual-state protocol.
Among methods covering all objects, it attains the lowest RS and AS dgIoU, dcDist and dCD, the lowest range and pivot errors and the highest CLIP.
DIPO attains the lowest AOR and Articulate-Anything the lowest LPIPS.
Retrieved part meshes rarely interpenetrate and carry clean textures.
Their range and pivot errors are still two to seven times ours.
A plausible layout does not recover how the observed object moves.
A simulation asset must reproduce that motion.

The multi-state block shows how strongly a state-wise static prior depends on observations.
Four additional states lower the AS dgIoU of FreeArt3D from 0.948 to 0.376 and its range error from 1.051 to 0.163.
TransArt3D reaches an AS dgIoU of 0.357 from the two endpoints.
It also improves all AS geometry metrics, AOR, pivot, CLIP and LPIPS over six-state FreeArt3D.
The pivot error falls from 0.134 to 0.077.
The generated transition thus supplies the intermediate evidence that FreeArt3D collects from extra observations.
In summary, these gains come from coupling the states through the video transition prior.
One joint must explain the whole transition, so per-state errors do not enter the parts and joint as false motion.

\noindent\textbf{Qualitative Results.}
Fig.~\ref{fig:qualitative_comparison} compares the reconstructions at both endpoints.
LARM produces incomplete geometry that blurs at the articulated endpoint.
Its part segmentation and joint location are also wrong.
Its feed-forward prediction is bounded by articulated training data.
Articulate-Anything retrieves a plausible asset that looks like another object.
Its parts come from other objects in the database.
Two-state FreeArt3D misses the interior that the articulation exposes.
Two states give its state-wise prior too little evidence for this region.
TransArt3D recovers the exposed interior, and both endpoint poses follow one joint.
The reconstruction thus remains a faithful articulated asset at every state between the two observations.

\begin{table}[t]
    \centering
    \fontsize{8.0}{8.8}\selectfont
    \setlength{\tabcolsep}{3pt}
    \resizebox{\columnwidth}{!}{%
    \begin{tabular}{@{}lcccc@{}}
        \toprule
        Variant & G-CLIP $\uparrow$ & G-LPIPS $\downarrow$ & M-PSNR $\uparrow$ & M-LPIPS $\downarrow$ \\
        \midrule
        (a) Single-Endpoint & 0.885 & 0.063 & 8.271 & 0.520 \\
        (b) Average pooling & 0.835 & 0.061 & 16.317 & 0.505 \\
        (c) w/o Routed Init. & 0.894 & 0.053 & 11.811 & 0.487 \\
        (d) w/o Artic. Routing & 0.901 & 0.053 & 14.426 & 0.524 \\
        (e) w/o Meas. Guidance & 0.907 & 0.054 & 8.787 & 0.404 \\
        (f) Full & \textbf{0.908} & \textbf{0.052} & \textbf{17.237} & \textbf{0.389} \\
        \bottomrule
    \end{tabular}}
    \caption{Canonical appearance ablations on 48 objects with fixed geometry and articulation. G metrics use the full endpoint images, and M metrics use the visible moving-part regions.}
    \label{tab:ablation_appearance}
\end{table}

\begin{figure}[t]
    \centering
    \setlength{\tabcolsep}{0.35pt}
    \newcommand{\acrop}[1]{\includegraphics[width=\dimexpr(\columnwidth-10\tabcolsep)/6\relax]{Figures/#1}}
    \newcommand{\aheading}[1]{\parbox[c]{\dimexpr(\columnwidth-10\tabcolsep)/6\relax}{\centering\small #1}}
    \begin{tabular}{@{}cccccc@{}}
        \aheading{(a)} &
        \aheading{(b)} &
        \aheading{(d)} &
        \aheading{(e)} &
        \aheading{Full} &
        \aheading{GT} \\[1pt]
        \acrop{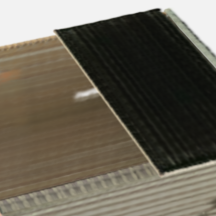} &
        \acrop{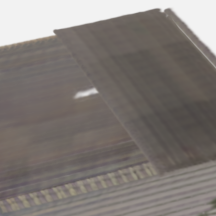} &
        \acrop{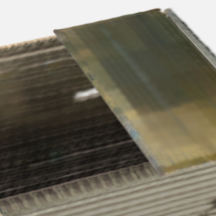} &
        \acrop{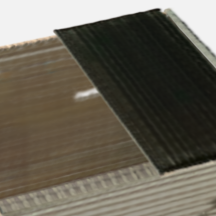} &
        \acrop{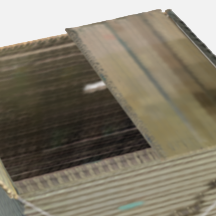} &
        \acrop{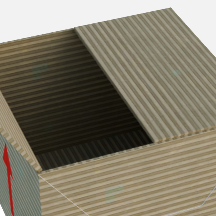}
    \end{tabular}
    \caption{Canonical appearance on a surface revealed only at the articulated endpoint.}
    \label{fig:appearance_ablation_viz}
\end{figure}

\subsection{Ablation Study}
\label{sec:ablation_study}

\noindent\textbf{Cross-state coupling.}
Removing cross-state coupling causes the largest drop in Tab.~\ref{tab:ablation_geometry}, and AOR nearly quadruples.
Each frame then keeps its own reconstruction errors, which the joint initialization reads as motion.
Consequently the parts interpenetrate (Fig.~\ref{fig:geometry_ablation_viz}), as for two-state FreeArt3D.
Endpoint correspondence initialization keeps geometry and range and nearly doubles the pivot error.
Two endpoints fix how far the part moves and leave the pivot loosely constrained.
Coupling the states is thus a prerequisite for estimating the articulation.

\noindent\textbf{Video guidance.}
Replacing video guidance with state-wise static 3D guidance worsens geometry, AOR and pivot even with six states.
More states improve each per-state fit.
They cannot reveal which differences between states are motion.
A coarse prompt such as ``open the drawer'' yields a transition that barely moves or moves wrongly, and the articulation follows it.
The video prior thus helps through the motion it depicts, once that motion is read out.
With the part assignment and joint fixed, video guidance alone barely helps.
Boundary refinement then cuts the range error by 40\%, and joint refinement cuts the pivot error by 70\%.
CHORD with a post-hoc fit reads the motion out after optimization and doubles the pivot error even on the ground-truth mesh.
Its free field absorbs the motion into non-rigid sliding (Fig.~\ref{fig:geometry_ablation_viz}).
The joint must therefore constrain the field during optimization.

\noindent\textbf{Canonical appearance.}
With geometry and articulation fixed, the variants in Tab.~\ref{tab:ablation_appearance} differ mainly on surfaces revealed only at the articulated endpoint (Fig.~\ref{fig:appearance_ablation_viz}).
A single endpoint never sees them, and average pooling blurs them.
Routing without the articulation places the observations on the wrong surface.
Without measurement guidance, the prior invents texture that keeps G-CLIP high and halves M-PSNR.
Faithful appearance thus depends on the recovered articulation, which decides where each observation belongs.

\subsection{Applications and Extensions}
\label{sec:application_extension}

\begin{figure}[!t]
    \centering
    \setlength{\tabcolsep}{0.5pt}
    \newcommand{\rim}[1]{\includegraphics[width=\dimexpr(\columnwidth-6\tabcolsep)/4\relax]{Figures/submission/#1}}
    \begin{tabular}{@{}cccc@{}}
        \multicolumn{2}{c}{\small Our Results} &
        \multicolumn{2}{c}{\small Articulation Parts}\\
        \scriptsize $t=0.5$ & \scriptsize $t=1$ &
        \scriptsize $t=0.5$ & \scriptsize $t=1$ \\[2\tabcolsep]
        \rim{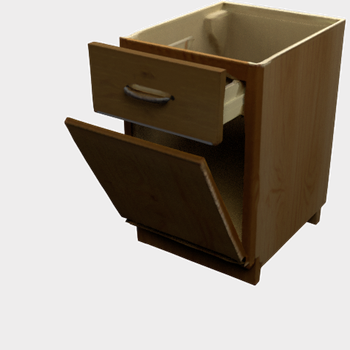} & \rim{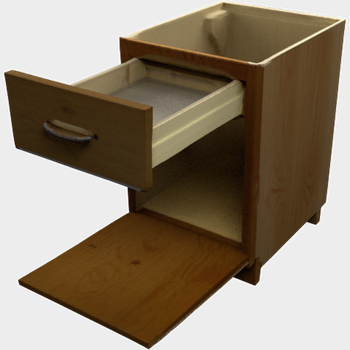} & \rim{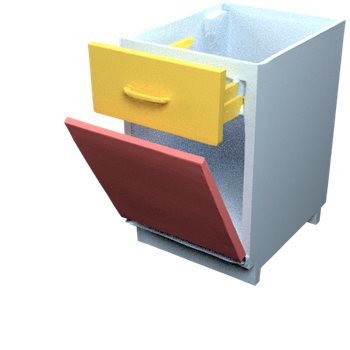} & \rim{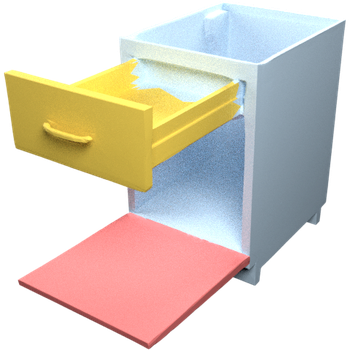}\\[\dimexpr2\tabcolsep-\dp\strutbox\relax]
        \rim{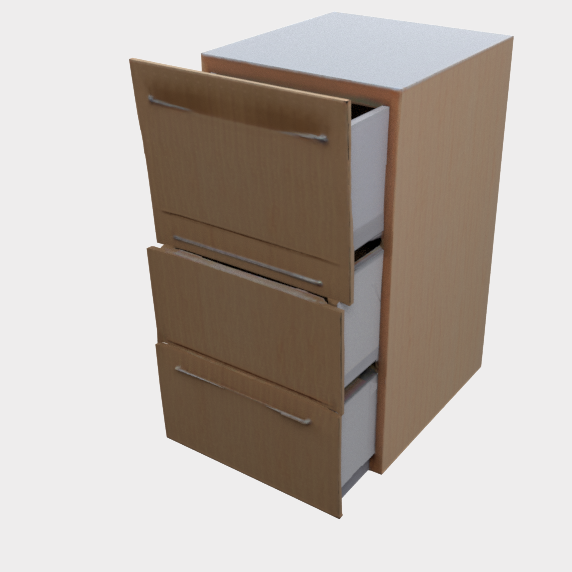} & \rim{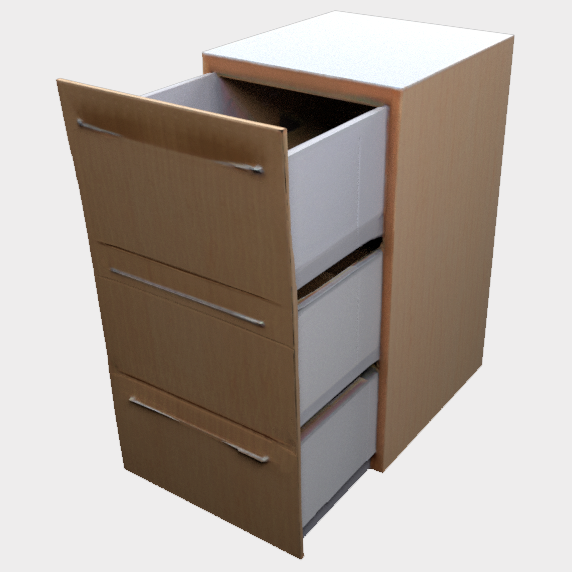} & \rim{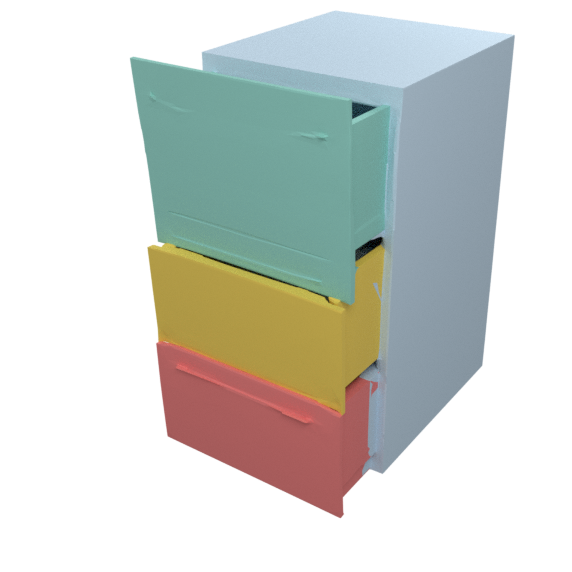} & \rim{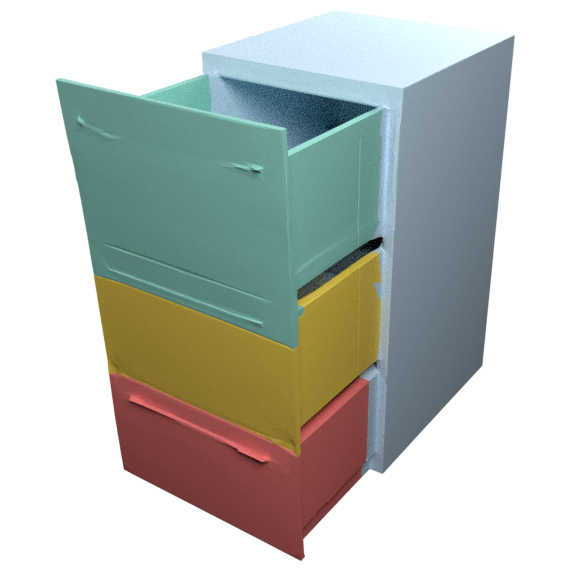}
    \end{tabular}
    \caption{Qualitative multi-joint extension showing reconstructed states and recovered movable-part decompositions.}
    \label{fig:multipart}
\end{figure}

\noindent\textbf{Multi-Joint Extension.}
For objects with several movable parts, each part has its own part probability and joint.
All of them constrain the shared motion field during refinement.
Fig.~\ref{fig:multipart} shows the recovered parts.

\begin{figure}[!t]
    \centering
    \setlength{\tabcolsep}{0.5pt}
    \newcommand{\rim}[1]{\includegraphics[width=\dimexpr(\columnwidth-6\tabcolsep)/4\relax]{Figures/#1}}
    \begin{tabular}{@{}cccc@{}}
        \multicolumn{2}{c}{\small Our Results} &
        \multicolumn{2}{c}{\small Inputs} \\
        \scriptsize $t=0.5$ & \scriptsize $t=1$ &
        \scriptsize $t=0$ & \scriptsize $t=1$ \\[2\tabcolsep]
        \rim{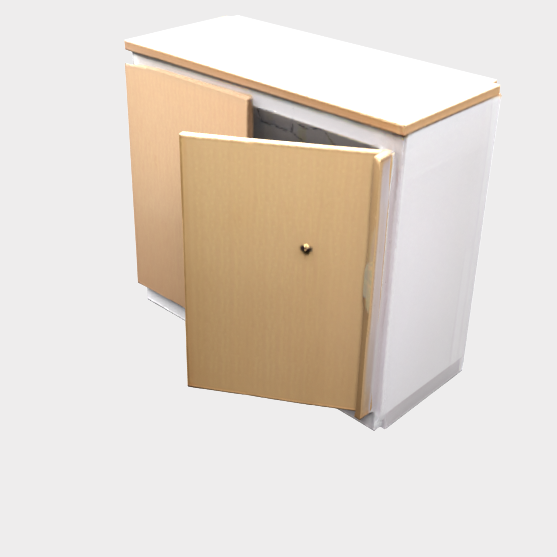} & \rim{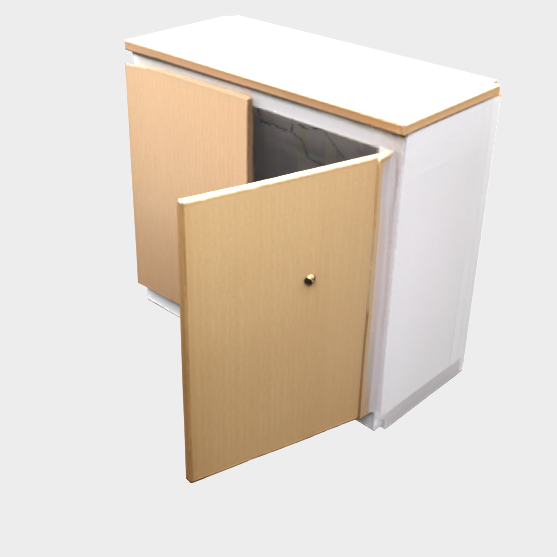} & \rim{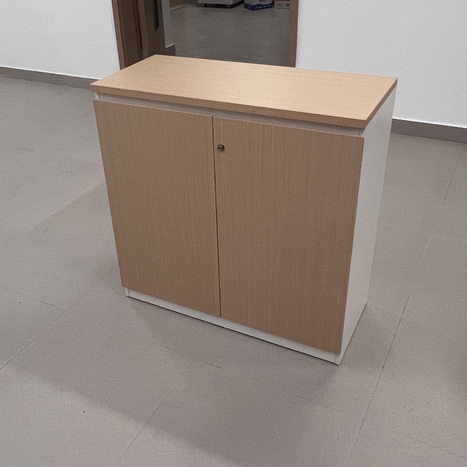}& \rim{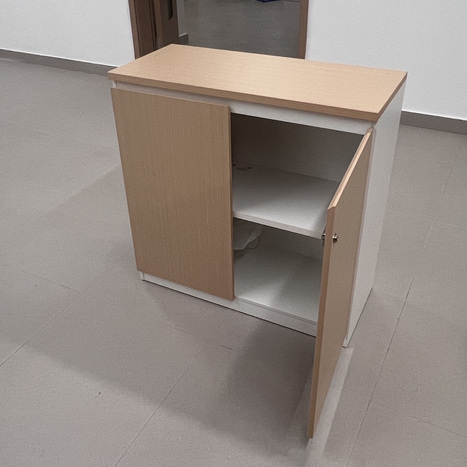}\\[\dimexpr2\tabcolsep-\dp\strutbox\relax]
        \rim{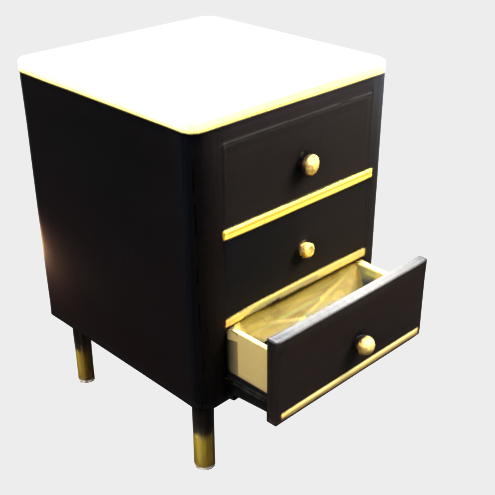} & \rim{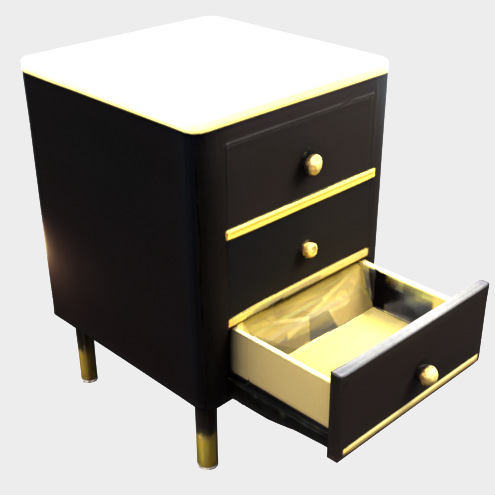} & \rim{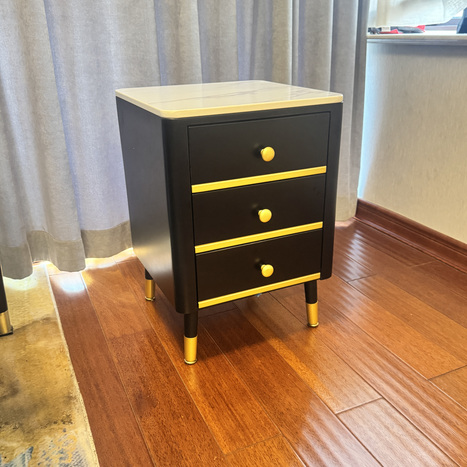}& \rim{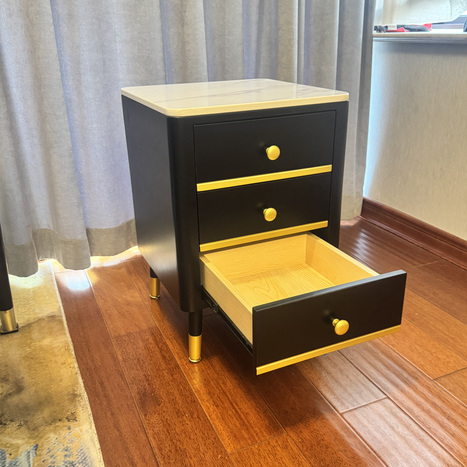}
    \end{tabular}
    \caption{Real-world articulated reconstruction from two casually captured endpoint images per object.}
    \label{fig:realworld}
\end{figure}

\noindent\textbf{Real-World Reconstruction.}
Fig.~\ref{fig:realworld} shows reconstructions from two casually captured endpoint photos per object.
TransArt3D recovers textured articulated assets that can be posed between the two observed states.
It needs no articulated training data and therefore applies directly to real photos.


\section{Conclusion}
\label{sec:conclusion}
We present TransArt3D, an optimization-based framework that reconstructs an articulated 3D asset with explicit parts and joint from two endpoint images. It translates a video transition prior into articulation by letting a part assignment and one shared joint constrain a video-supervised motion field over the whole transition, so per-state errors stay out of the joint. Coupling the states also initializes the parts and joint from jointly lifted frames and textures one canonical surface from both endpoints. Experiments show that TransArt3D recovers more accurate geometry, articulation and appearance than existing methods, and that two endpoint images with the transition prior surpass six independently observed states. It extends to real-world captures and multi-joint objects, turning two photographs of an object into an asset that a simulator can articulate.
\FloatBarrier

\noindent\textbf{Acknowledgements.}
This work is supported by the Science and Technology Development Fund of Macau Project (grant No. 0096/2023/RIA2, 0044/2024/AGJ, 0140/2024/AGJ, 0084/2024/RIB2), and is partially supported by the AI Supercomputing Center of Macau University of Science and Technology (MUST). We acknowledge the technical support provided by the Information Technology Development Office of MUST.

{
    \small
    \bibliographystyle{ieeenat_fullname}
    \bibliography{main}
}

\end{document}